\documentclass{article}

\pdfoutput=1

\usepackage{arxiv}

\usepackage[utf8]{inputenc} 
\usepackage[T1]{fontenc}    
\usepackage{hyperref}       
\usepackage{url}            
\usepackage{booktabs}       
\usepackage{amsfonts}       
\usepackage{nicefrac}       
\usepackage{microtype}      
\usepackage{lipsum}		
\usepackage{graphicx}
\usepackage{natbib}
\usepackage{doi}
\usepackage{amsmath}

\usepackage{bm}
\usepackage{multirow}

\title{Multi-agent Policy Optimization with Approximatively Synchronous Advantage Estimation}

\author{ \hspace{1mm}Lipeng Wan* \\
	Department of Artificial Intelligence\\
	Xi'an Jiaotong University\\
	\texttt{wanlipeng@stu.xjtu.edu.cn} \\
	\And
	\hspace{1mm}Xuwei Song* \\
	Department of Artificial Intelligence\\
	Xi'an Jiaotong University\\
	\texttt{songxw17@stu.xjtu.edu.cn} \\
	\And
	\hspace{1mm}Xuguang Lan \\
	Department of Artificial Intelligence\\
	Xi'an Jiaotong University\\
	\texttt{xglan@mail.xjtu.edu.cn} \\
	\And
	\hspace{1mm}Zeyang Liu \\
	Department of Artificial Intelligence\\
	Xi'an Jiaotong University\\
	\texttt{zeyang.liu@stu.xjtu.edu.cn} \\
	\And
	\hspace{1mm}Nanning Zheng \\
	Department of Artificial Intelligence\\
	Xi'an Jiaotong University\\
	\texttt{nnzheng@mail.xjtu.edu.cn} \\
}

\renewcommand{\shorttitle}{\textit{arXiv} Template}

\hypersetup{
pdftitle={A template for the arxiv style},
pdfsubject={q-bio.NC, q-bio.QM},
pdfauthor={David S.~Hippocampus, Elias D.~Striatum},
pdfkeywords={First keyword, Second keyword, More},
}

\begin{document}
\maketitle

\begin{abstract}
Current policy based multi-agent reinforcement learning methods introduce an asynchronous value function or advantage function for individual agent to achieve credit assignment. As a result, agents may be misled to update their policies for better cooperation with partners' outdated policies. In this work, we achieve credit assignment through the marginal advantage estimation, which is an expansion from single-agent advantage estimation to multi-agent systems. We further introduce approximations in multi-agent policy optimization for synchronous advantage estimation, and decompose the optimization problem into multiple sub-problems of single-agent situation. The experimental results on StarCraft multi-agent challenges and multi-agent particle environments demonstrate the superiority of our method to baselines with asynchronous estimation for credit assignment.
\end{abstract}

\section{Introduction}
In cooperative multi-agent reinforcement learning (MARL), each agent is treated as an independent decision-maker, but they can be trained together to learn cooperation. The common goal is to maximize the global return from the perspective of a team of agents. One of the challenges is multi-agent credit assignment \citep{Credit}: in cooperative settings, joint actions typically generate only global rewards, making it difficult for each agent to deduce its own contribution to the team's success. Credit assignment requires differentiated evaluation for each individual policy, but designing a specialized reward function for each agent is complicated and lacks generalization \citep{RewardShaping1,RewardShaping2}. 

To achieve credit assignment, the paradigm of centralised training with decentralised executions (CTDE) is proposed \citep{CTDE_2,CTDE_1}. The basic idea of CTDE is to train agents together with global information. In policy based MARL, CTDE is realised by introducing centralised critics. Such critics only work during training and are accessible to additional state information. At the same time, for decentralised executions, each agent is assigned with an independent policy which conditions on its local action-observation history. 

\begin{figure}[htb]
\begin{center}
\includegraphics[scale=0.26]{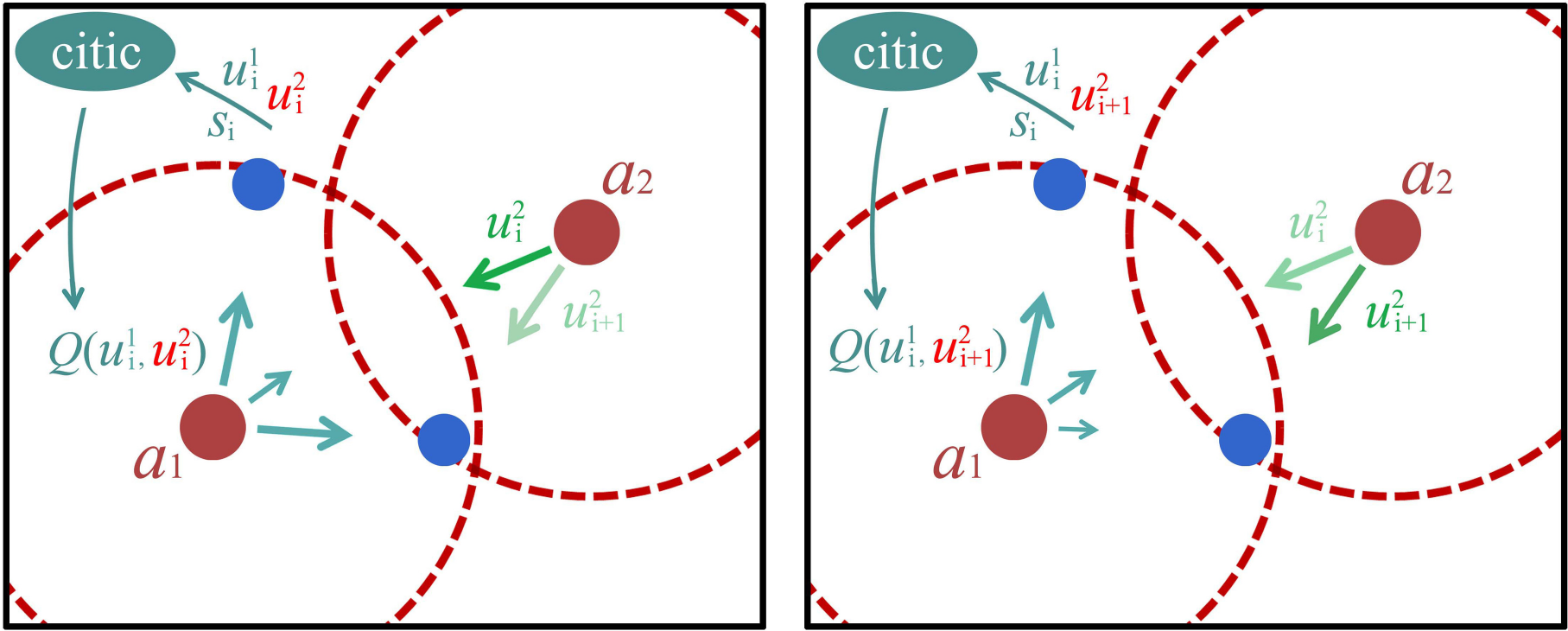}
\end{center}
\caption{Examples of asynchronous $Q$ value estimation (left) and synchronous $Q$ value estimation (right). Two agents ($a_1$, $a_2$) with limited sight range (red dashed circles) need to cover two landmarks (blue dots) while avoid collision. Estimated $Q$ values for the available actions of $a_1$ are represented by blueish arrows. The size of the arrow represents the scale of the $Q$ value. With asynchronous estimation, the policy of $a_1$ may be updated to move to the right, which results in collision with $a_2$.}
\end{figure}

In MARL, the centralised critic needs to estimate the policies of different agents respectively but simultaneously. Such estimations are based on the joint policy of all agents. For agent $a$, the estimation of its policy $\pi^a$ at iteration $i$ serves for the update from $\pi^a_i$ to $\pi^a_{i+1}$. Such update should be executed to optimize the joint policy $(\pi^a_{i+1},\pi^{-a}_{i+1})$, where $\pi^{-a}$ is the policy of the group of agents except $a$. The estimation of $(\pi^a_{i+1},\pi^{-a}_{i+1})$ is defined as synchronous estimation. Synchronous estimation requires the knowledge of other agents' future policy $\pi^{-a}_{i+1}$, as shown in Figure 1.

Current works such counter-factual multi-agent policy gradient (COMA) \citep{COMA} and multi-agent deep deterministic policy gradient \citep{MADDPG} achieve credit assignment with asynchronous estimation, where the action samples of other agents' current or past policies are considered. As a result, the policy of an individual agent is updated for cooperation with the current or past policies of their partners, which may eventually deteriorate the team's joint policy. For example, in COMA, based on the asynchronous estimation of $(\pi^a_i,\pi^{-a}_i)$, the policy of agent $a$ is updated according to the target: $Q(\pi^a_{i+1},\pi^{-a}_i) > Q(\pi^a_i,\pi^{-a}_i)$. However, the team's joint policy may degenerate as $Q(\pi^a_{i+1},\pi^{-a}_{i+1})<Q(\pi^a_i,\pi^{-a}_i)$. From the example, credit assignment with asynchronous estimation may introduce errors for the policy update. The problem is serious for situations where the policy updates frequently or dramatically in training. 

In this paper, we achieve credit assignment through the proposed marginal advantage estimation (MAE). Based on MAE, a novel method for approximatively synchronous advantage estimation (ASAE) in multi-agent policy optimization (MAPO) is proposed. Specifically, the advantage functions of single-agent reinforcement learning are extended to multi-agent systems for credit assignment. Then, the approximation of other agents' future policies is applied to achieve synchronous estimation. To ensure the reliability of the approximation, extra restrictions are introduced. The MAPO problem with introduced restrictions is further simplified and decomposed into multiple sub-problems, which are finally solved by proximal policy optimization method. Our algorithm is trained and tested on StarCraft multi-agent challenges and multi-agent particle environments. Experimental results show that our method outperforms baselines on most of the tasks. Compared to asynchronous estimation method, ASAE exhibits stronger ability to solve the challenge of credit assignment.

We have two contributions in this work: (1) A novel advantage estimation method which extends single-agent advantage functions to multi-agent system for credit assignment is proposed. (2) A simple yet effective method for synchronous policy estimation is proposed for the first time.

\section{Related Work}
A crucial challenge in cooperative multi-agent tasks is credit assignment. Reinforcement learning algorithms designed for single-agent tasks ignore credit assignment and show poor performance in complex cooperative tasks where high coordination among agents is required \citep{MADDPG}.

To deal with the challenge, some value based multi-agent reinforcement learning (MARL) methods introduce an utility function or local $Q$ value function to evaluate the policy for each agent. The global $Q$ value is then obtained from the local functions. In value decomposition network, global $Q$ value function is defined as the sum of local $Q$ value functions \citep{VDN}. Mixing Q network \citep{QMIX} acquires the global $Q$ value by mixing local $Q$ values with a neural network. In mean-field multi-agent methods, local $Q$ values are defined on agent pairs. The mapping from local $Q$ values to the global $Q$ value is established by measuring the influence of each agent pair's action to the shared return \citep{Mean_field}.

For policy based MARL methods, credit assignment is generally realised through differentiated evaluation with CTDE paradigm. Multi-agent deep deterministic policy gradient (MADDPG) extends DDPG to multi-agent tasks by introducing centralised critics. These critics are used to predict the $Q$ values for individual agent based on joint actions. Counter-factual multi-agent policy gradient (COMA) is inspired by the idea of difference reward \citep{Difference_reward} and provides a naive yet effective approach for differentiated advantage estimation in cooperative MARL. In COMA, a centralised critic is used to predict the $Q$ value function $Q(s,\textbf{u})$ for joint action $\textbf{u}$ and state $s$. And the advantage function for agent $a$ is defined as
\begin{equation}
A^a (s, \textbf{u}) = Q(s,\textbf{u}) - \sum_{u^a}{\pi^a(u^a|\tau^a) Q(s,(u^{-a},u^a))}
\end{equation}
where $\tau$ and $\pi$ represent trajectory and policy respectively. $-a$ denotes the group of agents except $a$, and $\textbf{u}=(u^a,u^{-a})$. However, COMA introduces a counter-factual baseline, which assumes that other agents take fixed actions sampled from current policies. The policy estimation for COMA is asynchronous. The problem is even more serious for MADDPG because the estimation is based on historical actions which are no longer taken by other agents.

\section{Background}
We consider a most general setting of partially observable, full cooperative multi-agent reinforcement learning (MARL) tasks, which can be described as a stochastic game defined by a tuple $G=<S,U,P,r,Z,O,n, \gamma>$. The true state of environment $s\in S$ is unavailable to all agents. At each time step, $n$ agents identified by $a\in A\ (A= \{1,2,\cdots,n\})$ receive their local observations $z^a\in Z$, and take actions $u^a \in U$ simultaneously. The joint observation $\textbf{Z}=Z^n$ is acquired by the observation function $O(s,a):S\times A \rightarrow \textbf{Z}$. The next state is determined by joint action $\textbf{u}\in \textbf{U}\ (\textbf{U}=U^n)$ and the transition function $P(s'|s, \textbf{u}):S\times \textbf{U} \times S \rightarrow [0,1]$. The reward function $r(s, \textbf{u}):S\times \textbf{U} \rightarrow \mathbb{R}$ is shared by all agents, so as the discounted return $G_t=\sum_{t+i}^\infty \gamma^t r_{t+i}$. $\gamma\in [0,1)$ is a discount factor. 

In policy based MARL with centralised training with decentralised executions, each agent is assigned with an independent policy $\pi^a(u^a|\tau^a)$, which is trained on its local trajectory $\tau^a$ consisting of historical observations and actions $\{(z^a_0,u^a_0),(u^a_1,z^a_1),\cdots \}$. Besides, action-state value function $Q^\pi (s,\textbf{u})$ and state value function $V^\pi (s)$ are usually introduced for policy evaluation. The advantage function is represented by $A^\pi (s,\textbf{u})=Q^\pi (s,\textbf{u})-V^\pi (s)$. And symbols in bold denote the joint variable of group agents. For brevity, $\pi$ is used to denote $\pi(u|\tau)$.

\begin{figure*}[htbp]
\begin{center}
\includegraphics[scale=0.34]{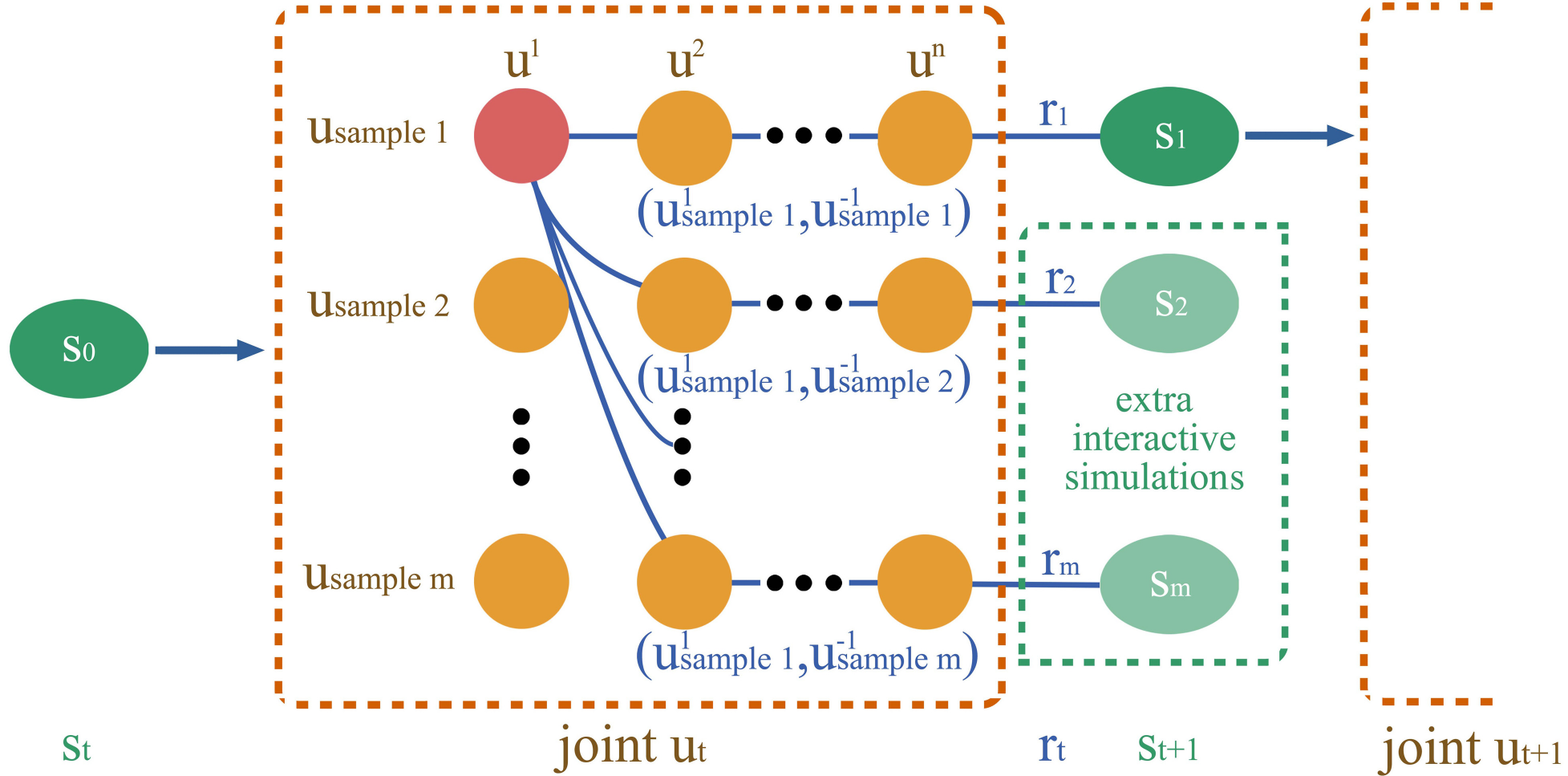}
\end{center}
\caption{Demonstration of MAE for action $u_{sample\ 1}^1$ (red dot). $m$ joint action samples are sampled at state $s_t$ (green ellipse). Each sample is represented by a lateral queue in orange dashed box. And the action of agent $n$ in sample $m$ is represented by $u_{sample\ m}^n$. These samples are then reorganized. Each blue solid line connects a reorganized sample. Based on $m$ reorganized samples, $m$ counter-factual advantages $A_{ctf}^1(u_{sample\ 1}^1,u_{sample\ 1}^{-1}), A_{ctf}^1(u_{sample\ 1}^1,u_{sample\ 2}^{-1}),\cdots,A_{ctf}^1(u_{sample\ 1}^1,u_{sample\ m}^{-1})$ are estimated according to equation (1) or equation (9). Especially, for counter-factual advantage function given by equation (9), extra interactive simulations (green dashed box) are executed. The marginal advantage is finally acquired by $A_{mar}^1 = \frac{1}{m} \sum_{i=1}^m A_{ctf}^1(u_{sample\ 1}^1,u_{sample\ i}^{-1})$.}
\end{figure*}

In single-agent policy optimization problems \citep{TRPO}, the objective is to maximize the expected accumulative reward, which is usually represented by an action-state value function. Similarly, for MAPO with centralised training with decentralised executions, agents are trained together with $Q$ values predicted by centralised critic. While each actor is expected to optimize a decentralised policy independently. The objective for agent $a$ is
\begin{equation}
\begin{aligned}
\max_{\pi^a} E_{u^a\sim\pi^a,u^{-a}\sim\pi^{-a}}\left[ Q^a(s,u^a,u^{-a}) \right]
\end{aligned}
\end{equation}
where $Q^a(s,u^a,u^{-a})$ is the estimated $Q$ value functions for agent $a$, which can be substituted by advantage functions to reduce the variance. The superscribe $-a$ denotes the group of $n-1$ agents except $a$.

\section{Approximatively Synchronous Advantage Estimation in Multi-agent System}
In this section, we first introduce marginal advantage estimation (MAE) which achieves credit assignment by extending advantage functions of single-agent reinforcement learning (RL) to multi-agent systems. Then, we describe how to implement synchronous advantage estimation based on MAE in multi-agent policy optimization (MAPO) problem.

\subsection{Credit Assignment through Marginal Advantage Estimation}
In this subsection, we address the challenge of credit assignment through the proposed marginal advantage estimation. 

From the perspective of teamed agents, the advantage function of joint action $\textbf{u}$ under state $s$ is defined as $A(s,\textbf{u})=Q(s,\textbf{u})-V(s)$. For clarity, we use $Q_{jt}$ to denote the $Q(s,\textbf{u})$. However, such advantage function is shared by the team. In order to implement differentiated advantage estimations for credit assignment, a particular marginal $Q$ value function $Q^a_{mar}$ is introduced for each agent.
\begin{equation}
Q^a_{mar}(s, u^{-a}) = E_{u^{-a}\sim\pi^{-a}} \left[ Q(s,\textbf{u})\right]
\end{equation}
where $\pi^{-a}(u^{-a}|\tau^{-a})$ is represented by $\pi^{-a}$ for brevity. $Q^a_{mar}$ is defined as the expectation of $Q_{jt}$ on other agents' policies $\pi^{-a}$, which is particular for each agent. Compared to $Q_{jt}$, $Q_{mar}$ also eliminates the estimation variance introduced by others' action samples. By replacing $Q_{jt}$ with $Q_{mar}$, the marginal advantage function $A^a_{mar}$ to achieve credit assignment is obtained:
\begin{equation}
\begin{aligned}
A^a_{mar} (s, u^a)=\int_{u^{-a}}Q(s,\textbf{u}) \ d\pi^{-a}- V(s)
\end{aligned}
\end{equation}

According to the Bellman equation, the state value function can be written as the expectation of $Q$ value function on policy. Since decentralised policies are considered to be independent \citep{COMA}, the joint policy $\bm{\pi}$ equals to a product of the independent policies where $\bm{\pi}=\prod_a^n\pi^a=\pi^a\cdot\pi^{-a}$.
\begin{equation}
\begin{aligned}
V (s) &= \int_\textbf{u} Q(s,\textbf{u}) \ d\bm{\pi} \\
&= \int_{u^{-a}} \int_{u^a} Q(s,u^a, u^{-a}) \ d\pi^a d\pi^{-a}
\end{aligned}
\end{equation}
Notice that $\textbf{u}=(u^a,u^{-a})$. Combine equation (4) with equation (5).
\begin{equation}
\begin{aligned}
A^a_{mar} (s, u^a)= \int_{u^{-a}} \left[ Q(s,\textbf{u}) - \int_{u^a} Q(s,\textbf{u}) \ d\pi^a \right] d\pi^{-a}
\end{aligned}
\end{equation}

Given a counter-factual scenario: only the actor under evaluation is considered to be agent, while other actors are viewed as part of the environment and always take fixed actions at the given state. The state in counter-factual scenario $s^a_{ctf}$ can be represented by  $(s, u^{-a})$. Apparently, $(s^a_{ctf}, u^a)=(s,u^a,u^{-a})=(s, \textbf{u})$. The advantage function under this circumstance is defined as counter-factual advantage function, which is written as:
\begin{equation}
\begin{aligned}
A^a_{ctf}(s^a_{ctf}, u^a) &= Q(s^a_{ctf},u^a) - V(s^a_{ctf}) \\
&= Q(s, \textbf{u}) - \int_{u^a} Q(s, \textbf{u}) \ d\pi^a
\end{aligned}
\end{equation}
Notice that $A^a_{ctf}$ is identical to the integral item in equation (6). By replacing the integral item with $A^a_{ctf}$, a simpler form of marginal advantage function is acquired.
\begin{equation}
\begin{aligned}
A^a_{mar} (s, u^a) =\int_{u^{-a}} A^a_{ctf}(s^a_{ctf}, u^a)\ d\pi^{-a}
\end{aligned}
\end{equation}
It can be proved that if $A^a_{ctf}$ is an unbiased estimation of $Q_{jt}$, and $A^a_{mar}$ is also an unbiased estimation of $Q^a_{mar}$ (Appendix I). 

Since the counter-factual scenario is defined as a single-agent environment, $A^a_{ctf}$ can be replaced by other forms of advantage function used in single-agent RL. For example, consider using TD residual $\delta_t^a$ as an estimation for $A^a_{ctf,t}$ at time step $t$.
\begin{equation}
\begin{aligned}
\delta_t^a &= r(s_{ctf},u^a_t)+\gamma V(s_{ctf,t+1})-V(s_{ctf, t}) \\
&=r(s_t, u^a_t, u^{-a}_t)+\gamma V(s_{t+1},u^{-a}_{t+1})-V(s_t,u^{-a}_t)
\end{aligned}
\end{equation}

Since agents' policies are considered to be independent \citep{COMA}, the integration on $\pi^{-a}$ can be split into a $(n-1)$-layer integration, which is complicated for execution. For simplicity and efficiency, the Monte-Carlo sampling is adopted.
\begin{equation}
\begin{aligned}
A^a_{mar}(s_t, u_t^a)&=\int_{u^{-a}} A^a_{ctf} (s_t, \textbf{u}_t)  d\pi^{-a}\\
&\approx \frac{1}{m} \sum^m_{i=1} A^a_{ctf} (s_t, u_t, u^{-a}_{t,i})
\end{aligned}
\end{equation}
Where $m$ is the number of joint action samples. For $m=1$ and $A^a_{ctf}$ given by equation (1), the marginal advantage function degenerates to advantage function of COMA. A demonstration of MAE with Monte-Carlo sampling is given in Figure 2.

\subsection{Synchronous Advantage Estimation through Approximations}
In this subsection, we describe how to achieve synchronous estimation in MAE through approximations. Synchronous policy estimation requires the prediction of other agents' future policies, as shown in Figure 1. Direct prediction of others' future policies is very difficult \citep{PolicyR}. Instead, in iterative training, only others' policies of the next iteration are needed. Assume others' joint policy of iteration $i$ is $\pi^{-a}_i$. The marginal advantage function with synchronous estimation can be written as:
\begin{equation}
A^a_{{i+1},syn} (s, u^a)=E_{u^{-a}\sim\pi_{i+1}^{-a}} \left[ A^a_{ctf, i+1}(s, u^a,u^{-a}) \right]
\end{equation}
Firstly, we introduce an approximation that $\pi_{i+1}^{-a} \approx \pi_i^{-a}$. To ensure the reliability, a restriction is introduced as $KL \left[ \pi^{-a}_{i+1}, \pi^{-a}_i \right] <\delta_1$. Besides, the advantage function is based
on trajectory samples, while only samples from iteration $i$ are available \citep{TRPO}. To ensure there is no much difference of trajectory distributions between episode $i$ and $i+1$, a supplementary restriction is introduced as $KL \left[ \pi^a_{i+1}, \pi^a_i \right] <\delta_2$. Combining the restrictions with equation (2), the objective for policy optimization of agent $a$ is
\begin{equation}
\begin{aligned}
&\max_{\pi_{i+1}^a}\ {E_{u^a\sim\pi^a_i}\left[ A^a_{i+1,syn} (s, u^a) \cdot\frac{\pi^a_{i+1}}{\pi^a_i} \right]} \\
&=\max_{\pi_{i+1}^a} {E_{\textbf{u}\sim \bm{\pi}^a_i} \left[ A^a_{i+1}(s, \textbf{u})\cdot \frac{\pi^a_{i+1}}{\pi^a_i} \right]} \\
&subject \ to: \ KL \left[ \pi^{-a}_{i+1}, \pi^{-a}_i \right] <\delta_1\\
&\ \ \ \ \ \ \ \ \ \ \ \ \ \ \ \ \ \ \ \ \ KL \left[ \pi^a_{i+1}, \pi^a_i \right] <\delta_2
\end{aligned}
\end{equation}
Notice that we have introduced importance sampling to  calculate the expectation on $\pi^a_{i+1}$ with the samples from $\pi^a_i$. The first restriction involves other agents' policies, which requires joint optimization of all agents' policies. The integral objective of multi-agent policy optimization with $n$ agents is
\begin{equation}
\begin{aligned}
&\max_{\pi_{i+1}^a} \sum_a^n {E_{\textbf{u}\sim \bm{\pi}_i} \left[ A^a_{i+1}(s, \textbf{u})\cdot \frac{\pi^a_{i+1}}{\pi^a_i} \right]} \\
&subject \ to: \bigcup^n_a \ KL \left[ \pi^{-a}_{i+1}, \pi^{-a}_i \right] <\delta_1\\
&\ \ \ \ \ \ \ \ \ \ \ \ \ \ \ \ \ \ \ \ \bigcup^n_a KL \left[ \pi^a_{i+1}, \pi^a_i \right] <\delta_2 \\
\end{aligned}
\end{equation}

It can be proved that $KL \left[ \pi^{-a}_{i+1}, \pi^{-a}_i \right] =\sum_o^{-a} KL \left[ \pi^o_{i+1}, \pi^o_i \right]$ (Appendix II). For simplification, a tighter form of the restriction $\ KL \left[ \pi^{-a}_{i+1}, \pi^{-a}_i \right] <\delta_1$ can be written as
\begin{equation}
KL \left[ \pi^o_{i+1}, \pi^o_i \right]< \frac{\delta_1}{n-1}=\delta_1' , \ for \ o \ in \ \bigcup^{-a}
\end{equation}
By replacing the restriction $\ KL \left[ \pi^{-a}_{i+1}, \pi^{-a}_i \right] <\delta_1$ with the tighter form, the restriction $\bigcup^n_a \ KL \left[ \pi^{-a}_{i+1}, \pi^{-a}_i \right] <\delta_1$ equals to
\begin{equation}
\bigcup^n_a \bigcup^{-a}_o \lbrace KL \left[ \pi^o_{i+1}, \pi^o_i \right]<\delta_1' \rbrace _a
\end{equation}
Notice that there are $n-1$ duplicate restrictions for each $ KL \left[ \pi^a_{i+1}, \pi^a_i \right]<\delta'$. By removing redundant duplicates, the restriction $\bigcup^n_a \ KL \left[ \pi^{-a}_{i+1}, \pi^{-a}_i \right] <\delta_1$ is simplified.
\begin{equation}
\bigcup^n_a  KL \left[ \pi^a_{i+1}, \pi^a_i \right]<\delta_1'
\end{equation}
Set $\delta_1'=\delta_2$ and two restrictions in equation (13) can be combined into $\bigcup^n_a  KL \left[ \pi^a_{i+1}, \pi^a_i \right]<\delta_2$.
In centralised training with decentralised executions, the policies of different agents are updated independently. For agent $a$, only the sub-restriction $KL \left[ \pi^a_{i+1}, \pi^a_i \right]<\delta_2$ is effective. For further simplification, the integral objective in equal (13) can be split into $n$ sub-objectives:
\begin{equation}
\begin{aligned}
for \ a \ in \ &{1,2,\cdots,n}:\\
&\max_{\pi_{i+1}^a} {E_{\textbf{u}\sim \bm{\pi}_i} \left[ A^a_{i+1}(s, \textbf{u})\cdot \frac{\pi^a_{i+1}}{\pi^a_i} \right]} \\
& subject\ to:\ KL \left[ \pi^a_{i+1}, \pi^a_i \right]<\delta_2
\end{aligned}
\end{equation}

\begin{figure*}[ht]
\begin{center}
\includegraphics[scale=0.65]{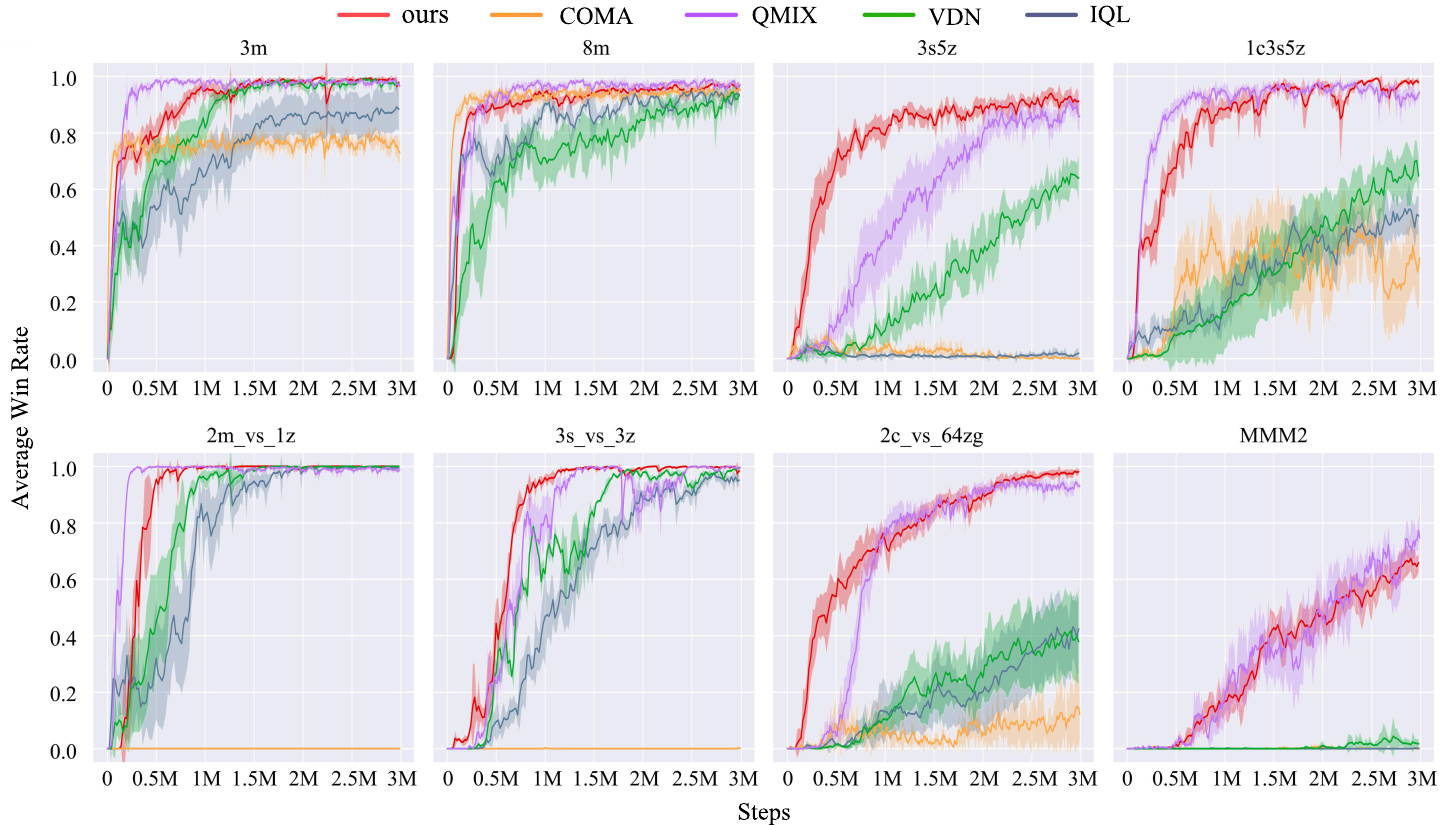}
\end{center}
\caption{Test win rate vs. training steps of various methods on SMAC benchmarks}
\end{figure*}

It is proved in the proximal policy optimization (PPO) that the KL divergence restriction can be effectively replaced by a clip operation \citep{PPO}. Referring to the PPO algorithm, the sub-objectives of MAPO with ASAE is acquired.
\begin{equation}
\begin{aligned}
&for \ a \ in \ {1,2,\cdots,n}:\\
&\max_{\pi_{i+1}^a} E_{u^a\sim\pi^a_i} \left[ E_{u^{-a}\sim\pi^{-a}_i}\left[ A^a_{i+1}(s, \textbf{u})\right]\cdot  \right.\\
&\left.\ \ \ \ \ \ \ \ \ \ \ \ \ \ \ \ \ \ \ \ \ \ \ \ \ \ \ clip(\frac{\pi^a_{i+1}}{\pi^a_i}, 1-\epsilon, 1+\epsilon) \right]
\end{aligned}
\end{equation}
Where $\epsilon$ is the clip range.

The training process of MAPO with ASAE is given by algorithm 1 (Appendix III).

\section{Experiments}
We evaluate our method (ASAE) on a variety of benchmarks. In experiments, the advantage function in equation (1) is applied as the counter-factual advantage for our method. We also adopt the CTDE paradigm, and use the same actors and critic structures as COMA \citep{COMA}. The centralised critic is used to predict the joint $Q$ values of reorganized samples. The number of samples $m$ is 10 and the clip range is 0.1 for all experiments.

\subsection{StarCraft Multi-agent Challenge (SMAC)}
\textbf{Environment Settings.} SMAC \citep{SMAC} is a benchmark designed for the evaluation of cooperative MARL algorithms \citep{grid,efficient}. It provides an interface for RL agents to interact with StarCraft II. In our experiments, only ally units are considered to be MARL agents. The environment is partially observable and each agent has a sight range. Only specific attributes of units in sight range are observable. And agents is accessible to the terrain features surrounding them. In addition, the last actions of ally units in sight range are also observable. The state includes information about all units on the map and it is only available in centralised training.

We consider different types of tasks involving both mixed and single type of agents. Specifically, our experiments are carried out on 8 tasks of different difficulty level, and the detail information is given in appendix IV. In homogeneous tasks, agents are of the same type. In symmetric scenarios, each army is composed of the same units, while the enemy army always outnumbers allied army in asymmetric scenarios. In micro-trick tasks, a higher-level of cooperation and specific micromanagement tricks is required to defeat the enemy.

\textbf{Training Settings.} We compare our method with baseline algorithms including IQL \citep{IQL}, VDN \citep{VDN}, QMIX \citep{QMIX} and COMA. In these baselines, only COMA is policy based but suffers from the asynchronous estimation. For (super) hard tasks such as $2c\_vs\_64zg$ and $MMM2$, good samples are added to replay buffer in initializing stage for early policy training. Each algorithm is trained for 3 million time steps. The training batch-size is 32 and the discount factor is 0.99. The learning rates for critic and actors are both 0.001.

\textbf{Experimental Results.} The test win rates during training are shown in Figure 3. Compared to baselines, our algorithm performs best in most of the tasks. From the results, suffering from asynchronous estimation, COMA shows no performance in 3 tasks($2m\_vs\_1z$, $3s\_vs\_3z$, $MMM2$). And our algorithm shows considerable superiority to COMA in 7 out of 8 tasks.

The algorithms after 3 million steps of training are tested in 100 battle games. The win rates are given in table 1. From table 1 and Figure 3, it is noticeable that compared to QMIX, although our algorithm converges slower in some tasks such as $3m$ and $2m\_vs\_1z$, the final performance of our algorithm is even better than QMIX. It is probably because in our method, the introduced restrictions have limited the policy update, which slows down the convergence.

\begin{table}[htb]
\caption{Test win rates after 3 million steps of training}
\label{tab:tab2}
\begin{center}
\begin{tabular}{cccccc}
\multirow{2}*{\bf ENV} &\multicolumn{5}{c}{\bf Algorithms} \\ ~ & ours & COMA & IQL & VDN & QMIX \\ \hline \\
3m & \textbf{1.0} & 0.81 & 0.91 & \textbf{1.0} & \textbf{1.0} \\
8m & \textbf{1.0} & 0.97 & 0.88 & 0.91 & \textbf{1.0}\\
3s5z & \textbf{0.95} & 0.0 & 0.0 & 0.65 & 0.93\\
1c3s5z & \textbf{1.0} & 0.4 & 0.54 & 0.68 & 0.95 \\
2m$\_$vs$\_$1z & \textbf{1.0} & 0.0 & \textbf{1.0} & \textbf{1.0} & \textbf{1.0} \\
3s$\_$vs$\_$3z & \textbf{1.0} & 0 & 0.96 & \textbf{1.0} & \textbf{1.0} \\
2c$\_$vs$\_$64zg & \textbf{0.97} & 0.15 & 0.38 & 0.41 & 0.94 \\
MMM2 & 0.78 & 0 & 0 & 0 & \textbf{0.81} \\ \hline
\end{tabular}
\end{center}
\end{table}

We render the battle process between agents trained by our algorithms and default AI. Some key frames are showed in Figure 4. Agents in red are ally units. In the first task $3s5z$, the cooperative agents learn to focus fire after training. After few rounds of crossfire, enemy group quickly lose the first unit. In the second task $MMM2$, besides focus fire, the cooperative agents also learn to adjust formation and use skill to avoid being destroyed. Particularly, in micro-trick tasks, cooperative agents learn to take advantage of map features and units' differences. As shown in the third sub-graph, in task $2c\_vs\_64zg$, only ally units are able to move across terrain. Taking advantage of this, ally units can attack enemy and move across the terrain when enemy approaching thus avoid being attacked.

\begin{figure}[ht]
\begin{center}
\includegraphics[scale=0.27]{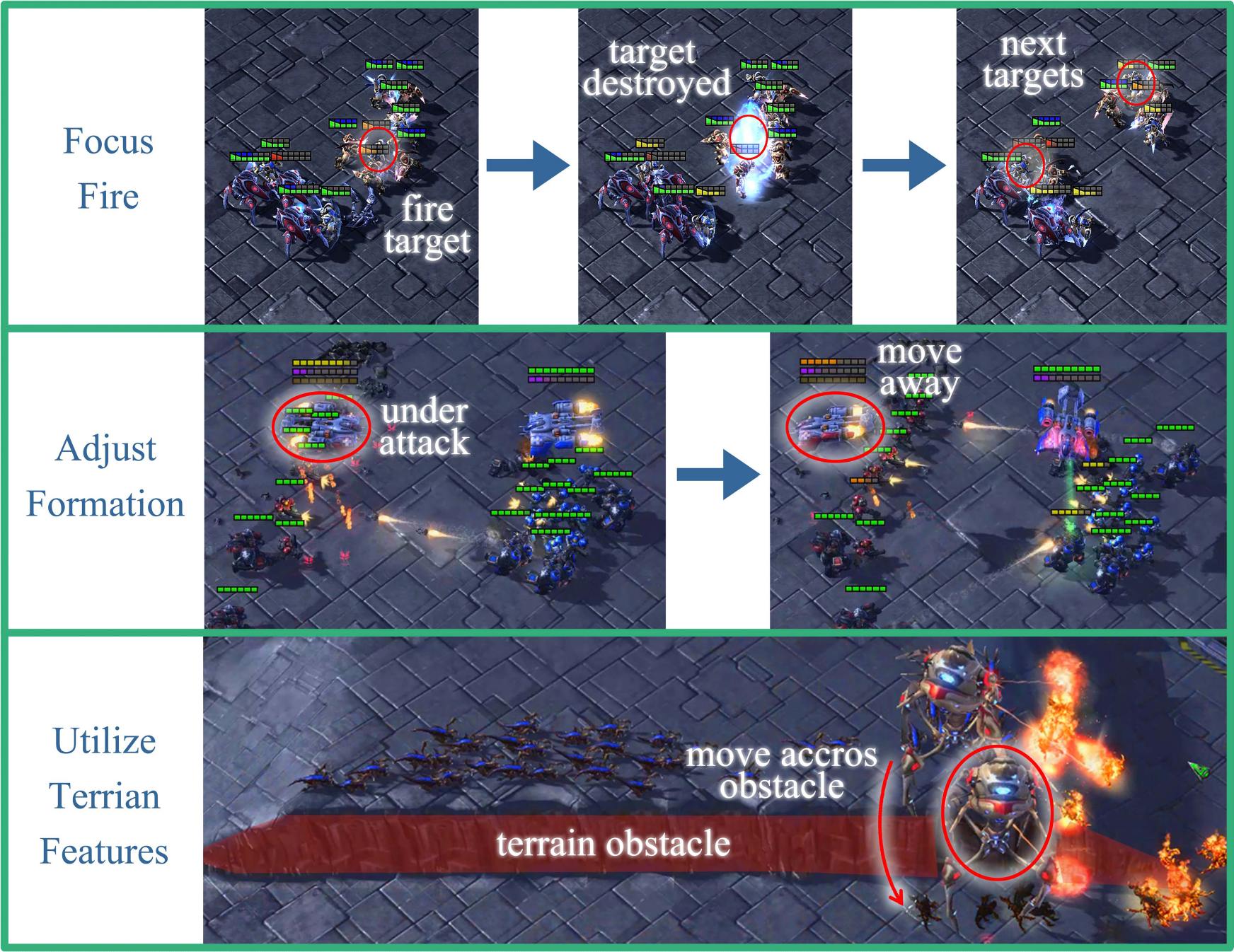}
\end{center}
\caption{Display of learned cooperative strategies.}
\end{figure}

\subsection{Simple Spread}
The simple spread environment from multi-agent particle environment\cite{particle} consists of $N$ agents and $N$ landmarks. The aim of teamed agents is to learn covering all the landmarks while avoiding collisions, as shown in Figure 5. At the beginning of each episode, the landmarks are randomly placed in a square box. The environment is partially observable, where each agent has a sight range. Only agents' speed, location and landmarks' location in sight range are observable. The reward function is designed based on the sum of distances from each agent to the nearest landmark. In addition, the team is punished when collision among agents occurs. The reward is shared by all agents. Our experiments are conducted in three independent tasks with $N=3$, $N=4$ and $N=5$.

\begin{figure}[tb]
\begin{center}
\includegraphics[scale=0.396]{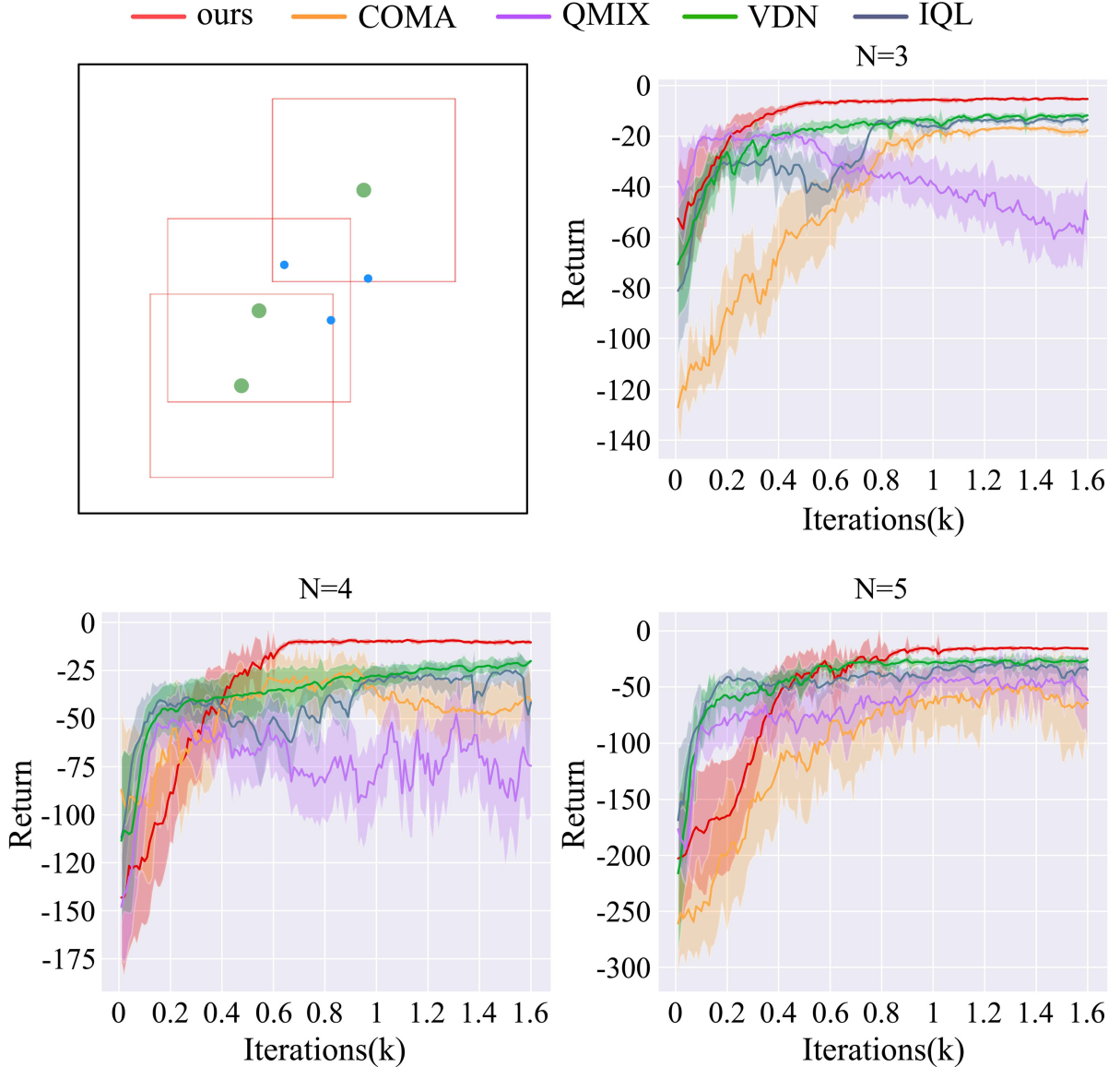}
\end{center}
\caption{Experiments on simple spread environment. The upper left sub-graph is a demonstration of the environment, where agents (green dots) need to cover all landmarks (blue dots) within finite sight range (red boxes) while avoid collision. Black dots denote the obstacles.}
\end{figure}

We compare our method with baseline algorithms including IQL, VDN, QMIX and COMA. Each algorithm is trained for 1.6k iterations. For each iteration, we train the network with all data collected from 4 environment threads. The max episode length for each thread is 150. The discount factor is 0.98 and learning rates for actor and critic are $5e-5$ and $5e-4$ respectively.

The training curves are shown in Figure 5. Our method converges after 0.8k iterations of training. Compared to baselines, our method achieves the highest return and stability.

\subsection{Predator-Prey}
The predator-prey environment from multi-agent particle environment \citep{particle} consists of obstacles, walls, predators and preys, as shown in Figure 6. The aim of predators is to capture all preys, while the aim of preys is to survive during a given time. Two stationary obstacles are randomly placed around the centre at the beginning of each episode. The environment is fully observable and agents is accessible to position of obstacles, walls, predators and preys. Besides, the speed of predators and preys are also observable. The predator team is rewarded based on the sum of the distances from each predator to the nearest prey. The predators also receive a positive reward when any prey is caught. The prey team is rewarded based on the sum of distances from each prey to the nearest predator, and is punished according to the sum of distance from each prey to the ground center. The prey team also receives a negative reward when any prey is captured. We conduct two different types of experiments involving both predators training for different algorithms and competitive training between our method and COMA.

\begin{figure}[tb]
\begin{center}
\includegraphics[scale=0.427]{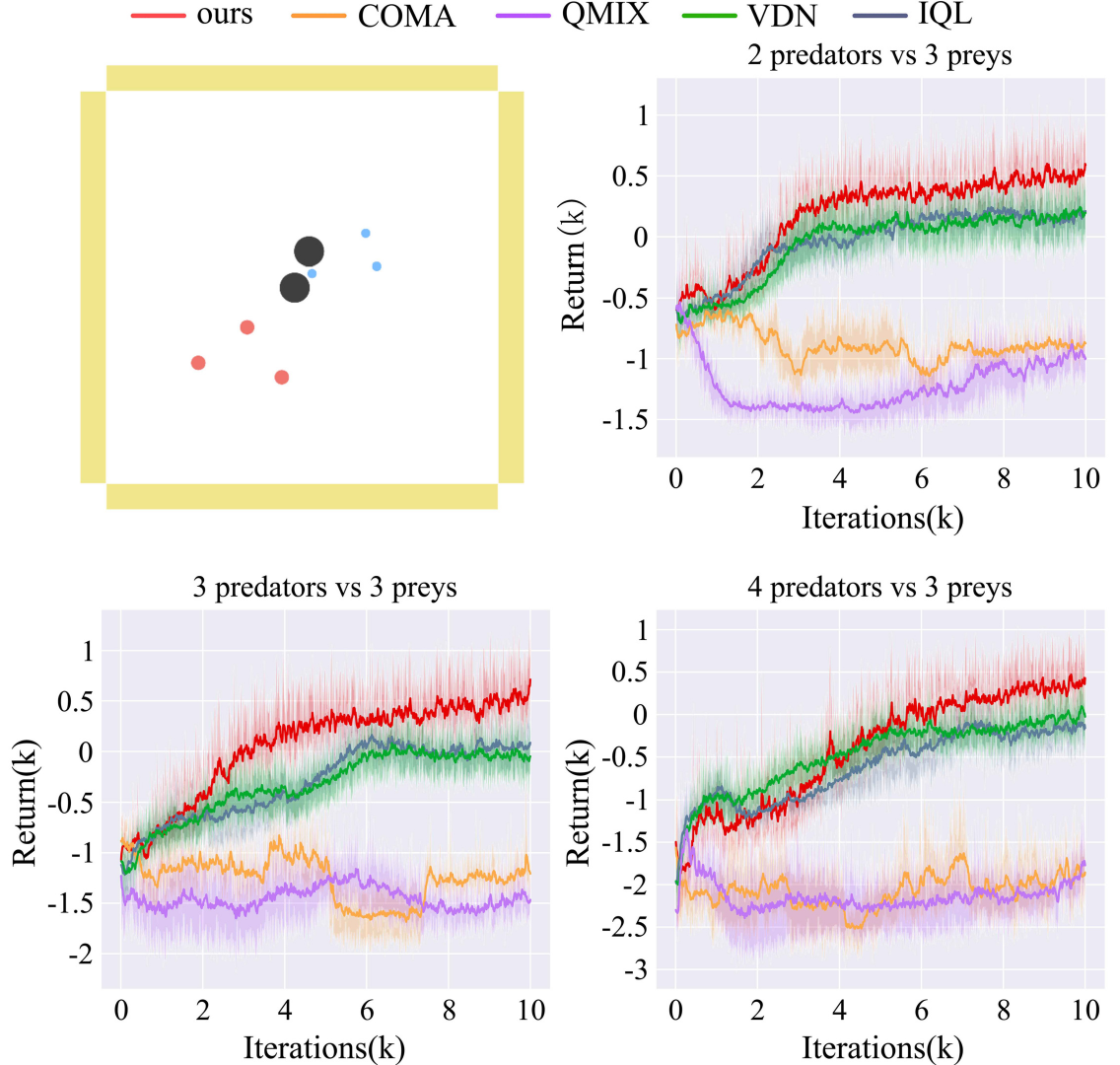}
\end{center}
\caption{Experiments on predators' training. The upper left sub-graph is a demonstration of the environment, where predators (red dots) hunt preys (blue dots) within ground surrounded by walls (yellow bars). Black circles denote the obstacles.}
\end{figure}

\begin{figure*}[htbp]
\begin{center}
\includegraphics[scale=0.47]{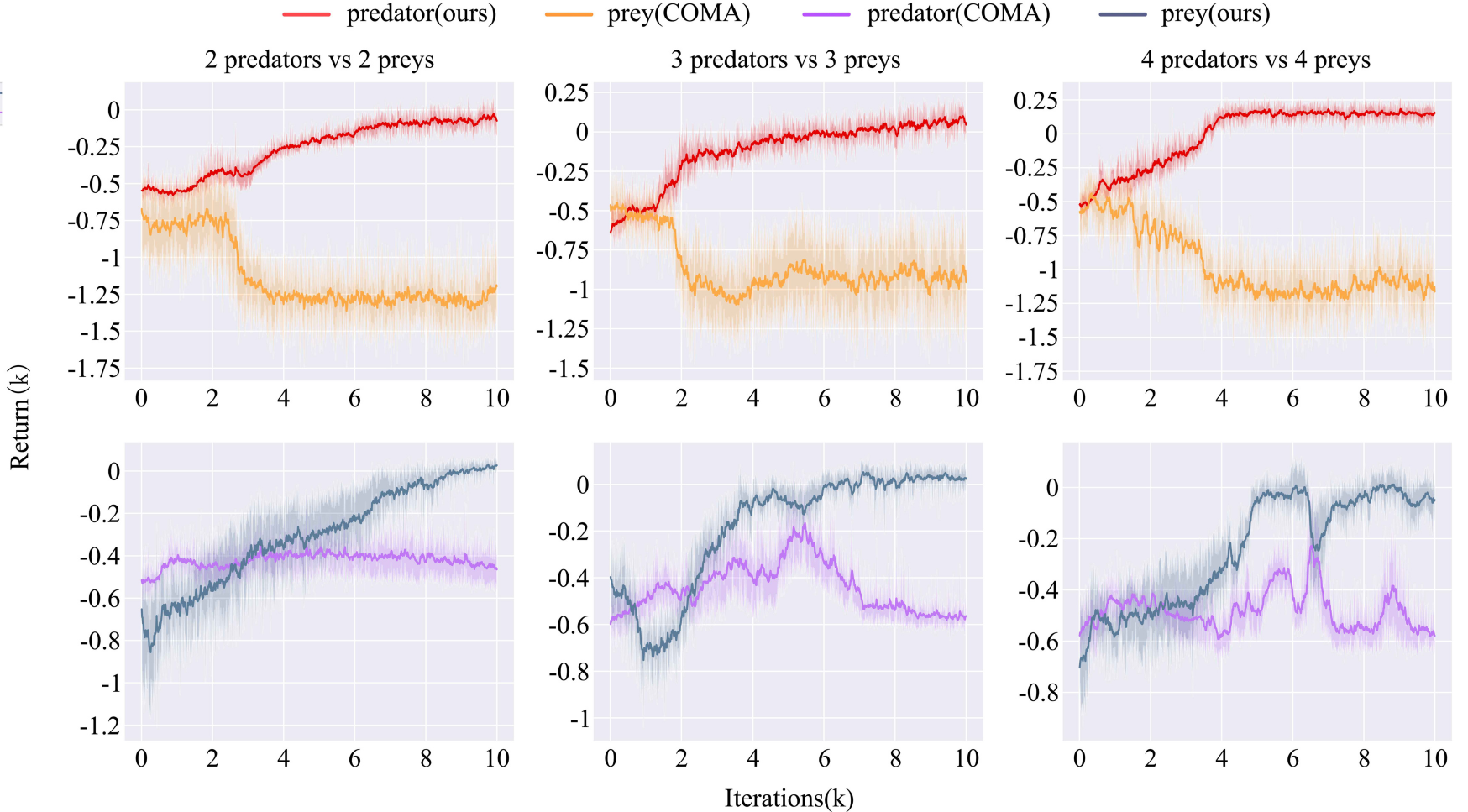}
\end{center}
\caption{Results of competitive training between our method and COMA.}
\end{figure*}

\textbf{Training of predators.} In experiment of predators' training, only predators are viewed as agents. While the preys are assigned with random policies. The number of prey is three. Each algorithm is trained for 10k iterations and other experiment settings are same as the experiments on simple spread. The experimental results are shown in Figure 6.

\textbf{Competitive training between ASAE and COMA.} We conduct experiments with competitive training for further comparison between our algorithm and COMA. Specifically, our method is adopted for training predator and prey group respectively, while COMA is always applied for the training of the opposite group. The numbers of predator and prey are the same in all tasks. 

The results are shown in Figure 7. In ASAE predators versus COMA preys scenarios, after 2k iterations of training, the performance of predators begins to improve with the training process, while the accumulative return of preys continues decreasing. Especially, in task $4\ predators\ vs\ 4\ preys$, the training curve of our method converges after 4k iterations training. For the opposite group setting, our method also shows better performance than COMA. We test the capture rate for predators after 10k iterations of competitive training. The capture rate is defined as the ratio of captured preys to total preys in 100 times of tests. The results are shown in Figure 8.

\begin{figure}[htbp]
\begin{center}
\includegraphics[scale=0.4]{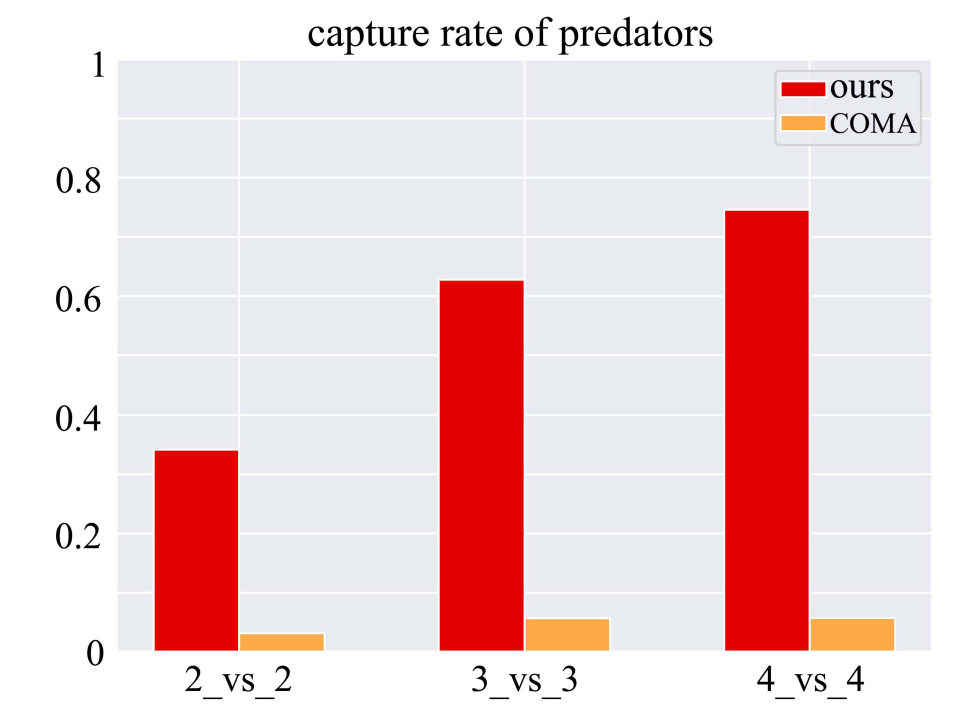}
\end{center}
\caption{Capture rate of trained predators.}
\end{figure}

\textbf{Discussion.} From the training curve of ASAE in Figure 5 and Figure 6, it can be seen that the credit assignment becomes more challenging as the number of cooperative agents increases. And it is noticeable that in Figure 7 and Figure 8, our algorithm shows growing superiority to COMA as the number of agent increases. From the results, it can be inferred that in comparison with COMA which applys asynchronous advantage estimations, our method is a better method to achieve credit assignment.

\section{Conclusion}
In this work, we describe the problem of asynchronous policy estimation for credit assignment in policy based cooperative MARL, and propose a approximatively synchronous estimation method to solve the problem. Firstly, we introduce the marginal advantage function which achieves credit assignment by extending single-agent advantage function to multi-agent systems. Then, the marginal advantage function is applied to MAPO, and the approximations are introduced to realise the synchronous estimation. We solve the optimization problem according to PPO. Our algorithms are evaluated on the StarCraft II SMAC and multi-agent particle environment. The experimental results demonstrates our algorithm outperforms the baselines in most of the tasks. Especially, compared to COMA, which suffers from asynchronous policy estimation, our algorithm shows increasing superiority in tasks with harder credit assignment challenge. 

ASAE achieves synchronous estimation through the approximations of parters' policies. For future work, we are interested in combining ASAE with policy modelling methods \citep{DBLP,han2018}, where partners' policies can be directly modelled.

\bibliography{SMAE}
\bibliographystyle{SMAE}

\appendix
\section{Appendix I}
Assume that joint advantage $A^a (s,\textbf{u})$ is a unbiased estimation of joint Q value function $Q(s,\textbf{u})$. Then
\begin{equation}
\begin{aligned}
A^a (s,\textbf{u})=Q&(s,\textbf{u})-b_s\\
where: \nabla_\theta& b_s \equiv0
\end{aligned}
\end{equation}
Then
\begin{equation}
\begin{aligned}
A^a (s,u)&=E_\pi^{-a} \left[ A^a(s,\textbf{u} ) \right]\\
&=E_\pi^{-a} \left[ Q(s,\textbf{u})-b_s \right] \\
&=E_\pi^{-a} \left[ Q(s,\textbf{u})\right]- E_\pi^{-a}\left[b_s \right] \\
\end{aligned}
\end{equation}
$E_\pi^{-a} \left[ Q(s,\textbf{u})\right]$ is exact the marginal Q value function and $\nabla_\theta E_\pi^{-a}\left[b_s \right]=E_\pi^{-a}\left[\nabla_\theta b_s \right] \equiv0$

\section{Appendix II}
Consider two agents, whose policies of episode $i$ are represented by $\pi_i^1$ and $\pi_i^2$ respectively.
\begin{equation}
\begin{aligned}
KL \left[ \pi_i^1 \pi_i^2, \pi^1_{i-1} \pi^2_{i-1}\right]&=\int \pi_i^1 \pi_i^2 \log \frac{\pi_i^1 \pi_i^2}{\pi_{i-1}^1 \pi_{i-1}^2}\ du\\
=& \int \pi_i^1 \pi_i^2 \left( \log \frac{\pi_i^1}{\pi_{i-1}^1}+\log \frac{\pi_i^2}{ \pi_{i-1}^2} \right) \ du\\
<& \int \pi_i^1 \log \frac{\pi_i^1}{\pi_{i-1}^1} \ du+ \int \pi_i^2 \log \frac{\pi_i^2}{ \pi_{i-1}^2} \ du\\
=&KL \left[ \pi_i^1, \pi^1_{i-1}\right] + KL \left[ \pi_i^2, \pi^2_{i-1}\right]
\end{aligned}
\end{equation}
The relation can be expanded to joint distribution of other agents' policies 
\begin{equation}
\begin{aligned}
KL \left[ \pi_i^{-a}, \pi^{-a}_{i-1}\right]&=\int \prod_o^{-a} \pi_i^o \log \frac{\prod_o^{-a} \pi_i^o }{\prod_o^{-a} \pi_{i-1}^o }\ du\\
&< \sum_o^{-a} KL \left[ \pi_i^o, \pi^o_{i-1}\right]
\end{aligned}
\end{equation}

\section{Appendix III}
Other experiments

\begin{figure}[htb]
\begin{center}
\includegraphics[scale=0.15]{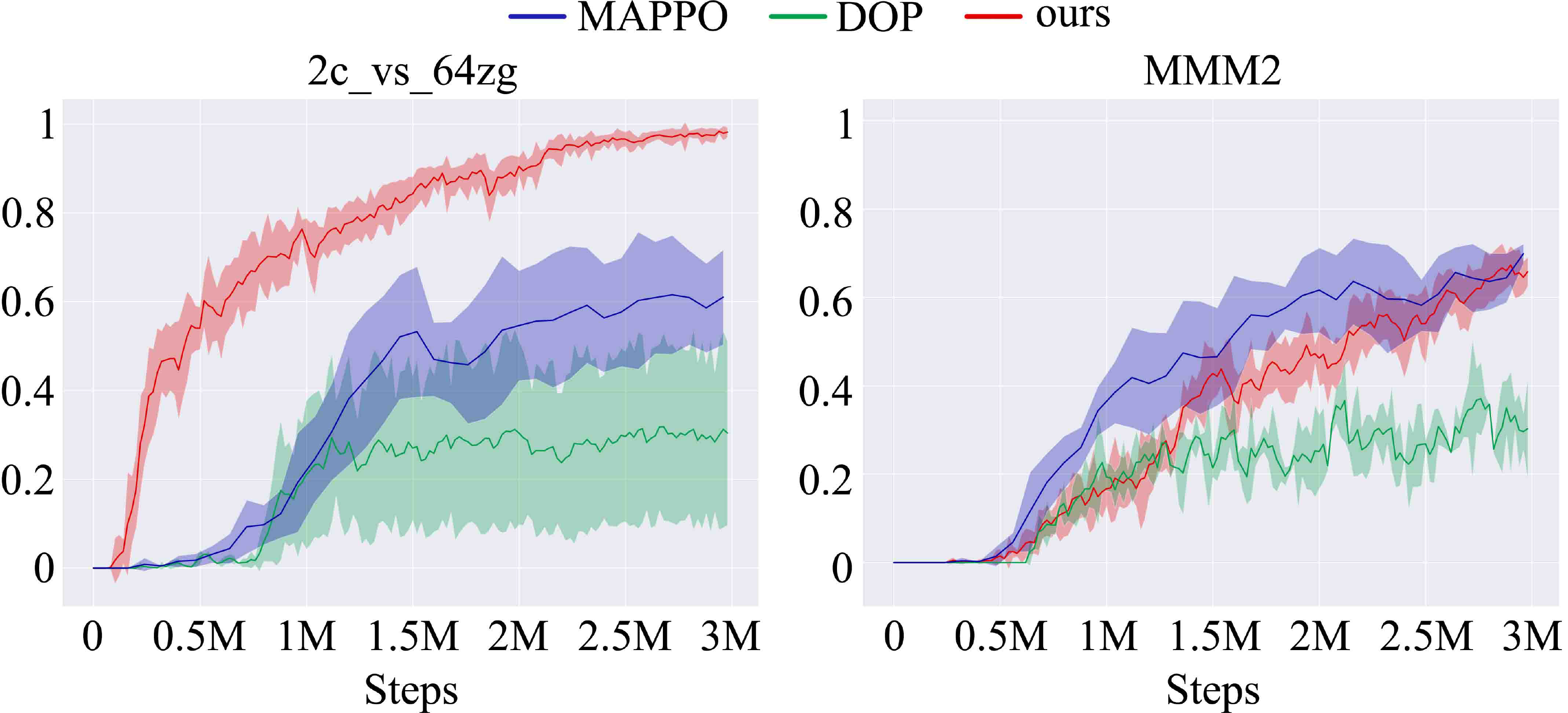}
\end{center}
\caption{Performance of our method with different number of action samples.}
\end{figure}

\begin{figure}[htb]
\begin{center}
\includegraphics[scale=0.15]{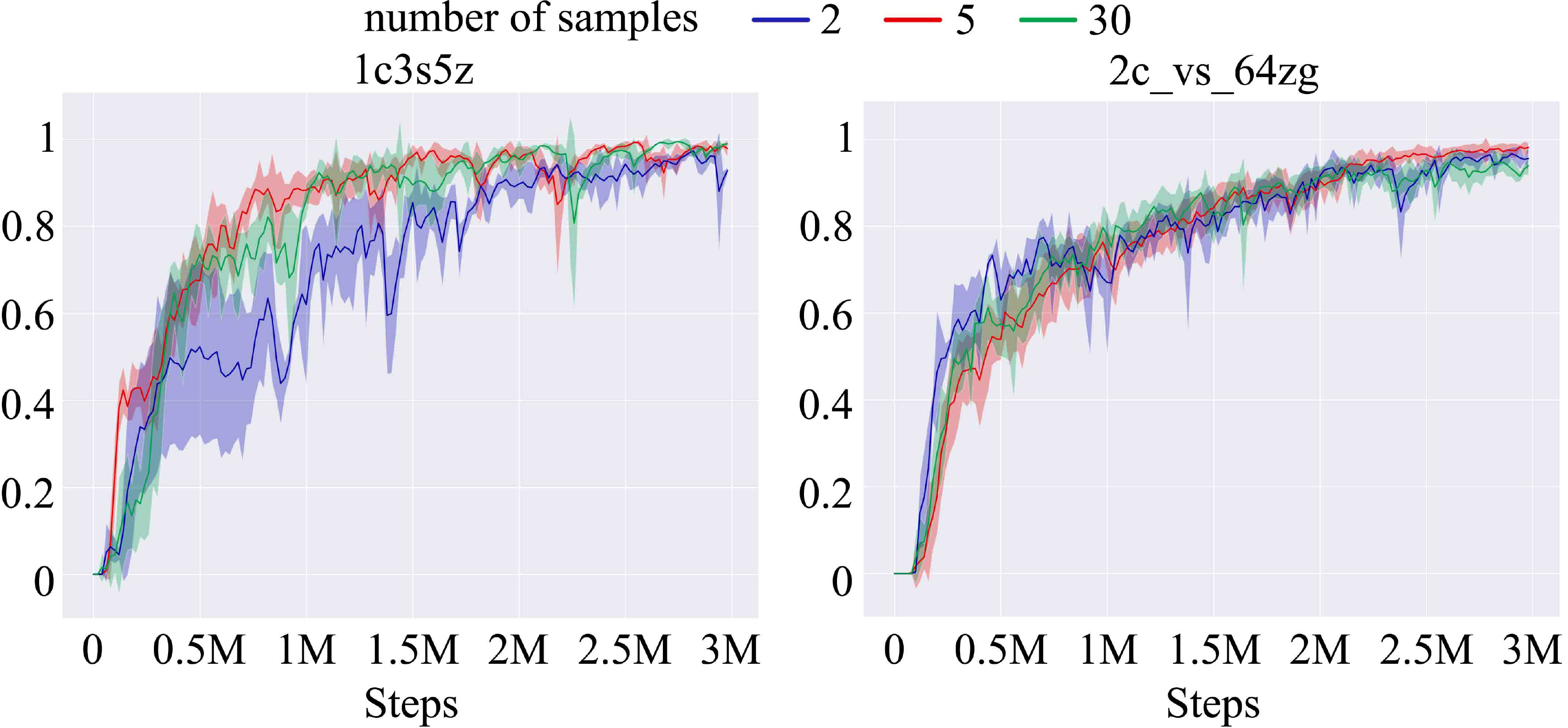}
\end{center}
\caption{Comparison among MAPPO, DOP and our method on SMAC.}
\end{figure}

\begin{figure}[htb]
\begin{center}
\includegraphics[scale=0.15]{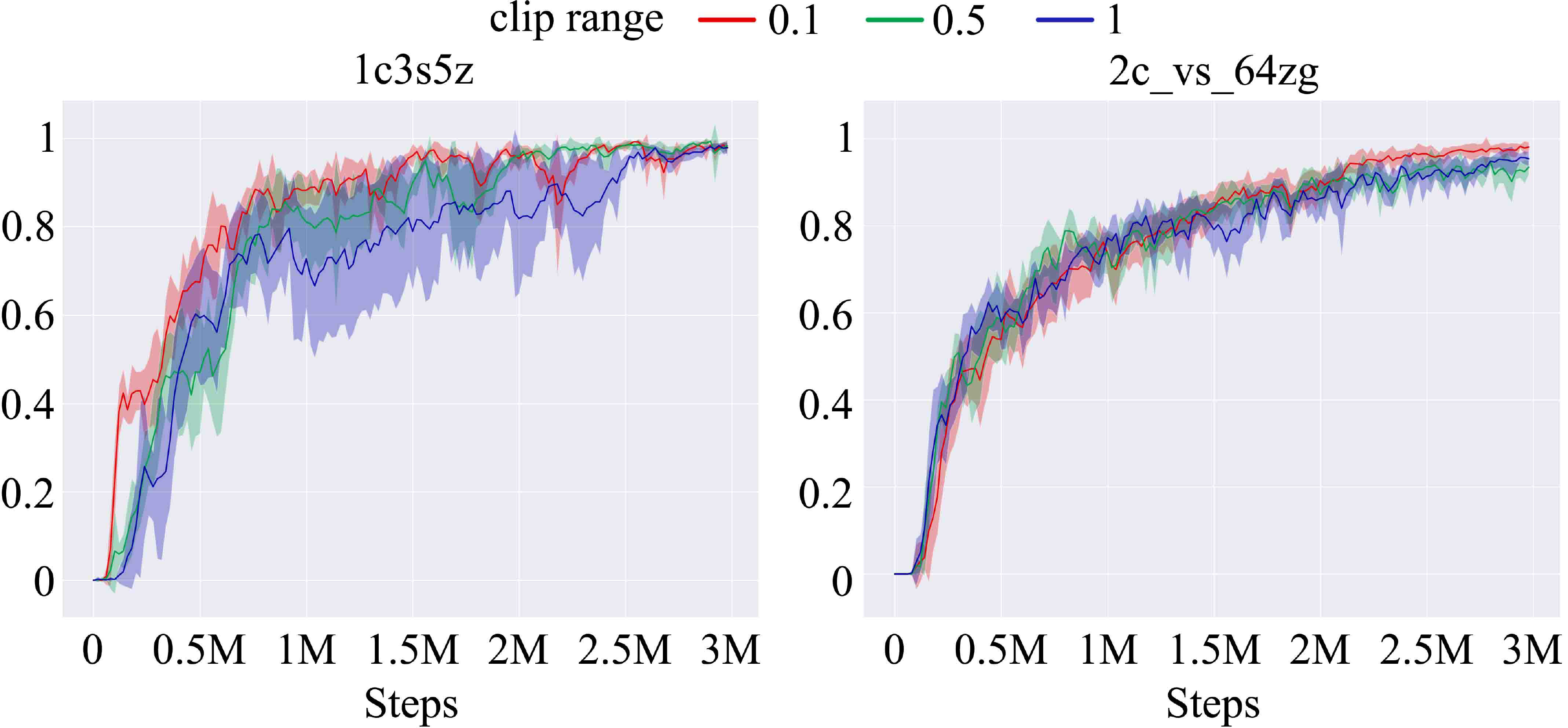}
\end{center}
\caption{Performance of our method with different clip ranges.}
\end{figure}

\end{document}